\documentclass[conference]{IEEEtran}
\usepackage{amsmath,amssymb,amsfonts}
\usepackage{algorithmic}
\usepackage{graphicx}
\usepackage{textcomp}
\usepackage{xcolor}
\usepackage{orcidlink}
\usepackage[inkscapelatex=false]{svg}
\usepackage[sorting=none, style=ieee]{biblatex}
\def\BibTeX{{\rm B\kern-.05em{\sc i\kern-.025em b}\kern-.08em
    T\kern-.1667em\lower.7ex\hbox{E}\kern-.125emX}}
\begin{document}

\title{
% A Deep Learning Tool for Touchscreen Authentication using Finger-Drawn Symbols\\
% AI-Powered Finger-Drawn Biometric Authentication\\
Neural Network-Powered Finger-Drawn\\Biometric Authentication\\
% A Finger-Drawn Biometric Authentication Mechanism\\
% \thanks{The project on which this report is based was funded by the Federal Ministry of Research, Technology and Space under the funding code “KI-Servicezentrum Berlin-Brandenburg” 16IS22092.}
}

\author{
\IEEEauthorblockN{1\textsuperscript{st} Maan Al Balkhi}
\IEEEauthorblockA{\textit{Freie Universität Berlin} \\
    Berlin, Germany \\
    \orcidlink{0009-0004-0585-4918}0009-0004-0585-4918
}

\and

\IEEEauthorblockN{2\textsuperscript{nd} Kordian Gontarska}
\IEEEauthorblockA{\textit{Hasso Plattner Institute} \\
    Potsdam, Germany \\
    \orcidlink{0000-0002-2755-405X}0000-0002-2755-405X
}

\and

\IEEEauthorblockN{3\textsuperscript{rd} Marko Harasic}
\IEEEauthorblockA{\textit{Fraunhofer FOKUS} \\
    Berlin, Germany \\
    \orcidlink{0000-0002-7210-893X}0000-0002-7210-893X
}

\and

\IEEEauthorblockN{4\textsuperscript{th} Adrian Paschke}
\IEEEauthorblockA{\textit{Fraunhofer FOKUS} \\
    Berlin, Germany \\
    \orcidlink{0000-0003-3156-9040}0000-0003-3156-9040
}
}

\maketitle

\begin{abstract}
This paper investigates neural network-based biometric authentication using finger-drawn digits on touchscreen devices. We evaluated CNN and autoencoder architectures for user authentication through simple digit patterns (0-9) traced with finger input.
Twenty participants contributed 2,000 finger-drawn digits each on personal touchscreen devices. We compared two CNN architectures: a modified Inception-V1 network and a lightweight shallow CNN for mobile environments. Additionally, we examined Convolutional and Fully Connected autoencoders for anomaly detection.
Both CNN architectures achieved $\sim$89\% authentication accuracy, with the shallow CNN requiring fewer parameters. Autoencoder approaches achieved $\sim$75\% accuracy.
The results demonstrate that finger-drawn symbol authentication provides a viable, secure, and user-friendly biometric solution for touchscreen devices. This approach can be integrated with existing pattern-based authentication methods to create multi-layered security systems for mobile applications.
\end{abstract}


\section{Introduction}
Protecting private and sensitive data on smartphones and similar devices is increasingly important as these devices store personal information, including bank account access data, social media profiles, and private contacts. Users access their mobile devices multiple times daily, making both security and usability crucial factors in authentication mechanisms. Traditional authentication methods like PINs or passwords can be easily copied or stolen \cite{biometric_authentication}, while even biometric methods such as fingerprint or iris recognition have been shown to be reconstructed or synthesized \cite{fingerprint_attack, iris_synthetizing}. Handwriting or finger-writing, as dynamic biometric features, cannot be easily reproduced by another person, offering a promising alternative for secure yet simple authentication \cite{rehman2019writer}.

The potential of drawing simple symbols and patterns with a finger for authentication purposes has been scarcely explored. While machine learning approaches have successfully matched handwritten texts or signatures with their authors \cite{rehman2019writer}, few studies have examined finger writing on touchscreens. Furthermore, most existing studies on signature verification use complex neural network architectures that may not be suitable for resource-constrained mobile devices \cite{rehman2019writer}. The authentication potential of simple symbols such as digits, which are easy to draw and remember, has not been thoroughly investigated.

This study aims to address this gap and contribute to the field of biometric authentication by determining whether neural networks can distinguish and authenticate people who draw simple symbols with their finger on touchscreens.
Specifically, we investigate:
\begin{itemize}
    \item Demonstrating the viability of finger-drawn simple symbols for person authentication
    \item Proposing a shallow CNN architecture that achieves high accuracy with fewer parameters than existing approaches
    \item Exploring autoencoder architectures as a novel approach to authentication as anomaly detection
    \item Providing insights into optimal parameters (image size, line width, kernel size) for neural network-based finger writing authentication
    % \item Analyzing the impact of device heterogeneity and symbol complexity on authentication accuracy 
\end{itemize}

The remainder of this paper is structured as follows. Section \ref{sec:rel_work} reviews the related work. Section \ref{sec:methodology} explains our methodology for the following evaluation. Section \ref{sec:results} shows our results. Section \ref{sec:discussion} discusses our results and concludes the paper.

\section{Related Work}\label{sec:rel_work}

\subsection{Finger Recognition Methods}

Previous research on finger-based authentication has primarily focused on fingerprint recognition. However, touchscreen interaction patterns have also been explored as behavioral biometric markers. Volaka et al. \cite{volaka2019towards} applied deep learning models to authenticate users based on touchscreen finger movement characteristics and sensor data, achieving 88\% accuracy and 15\% Equal Error Rate (EER) using the HMOG dataset \cite{yang2014multimodal}.

\subsection{Handwriting and Finger Writing Authentication Systems}

Takahashi et al. \cite{takahashi2021smartphone} investigated person verification through simple symbols on smartphones by extracting 40 individual features including coordinate features, maximum velocity, and symbol center. Using data from 30 participants drawing three symbols twenty times each (1800 total samples) on a single device, they achieved an EER of 10.6\%.

\begin{table}[htbp]
\caption{Achieved accuracies on different datasets by Mohapatra et al.}
\begin{center}
\begin{tabular}{|c|c|}
\hline
\textbf{Dataset} & \textbf{ACC} \\
\hline
 CEDAR & 100\%  \\
 \hline
 BHSig260-Bengali & 97,77\%  \\ 
 \hline
 BHSig260-Hindi & 95,40\%  \\ 
 \hline
 UTSig & 80,44\% \\
\hline
\end{tabular}
\label{tab1}
\end{center}
\end{table}

\subsection{Neural Network Approaches for Authentication}

For signature verification, Alajrami et al. \cite{alajrami2020handwritten} constructed a deep CNN to verify scanned signatures, classifying them as genuine or forged. Using a dataset of 30 people with five genuine and five forged signatures per person, they achieved 99.7\% accuracy with an 80:20 training-validation split.

Mohapatra et al. \cite{8985925} modified the Inception-V1 architecture for person verification using scanned signatures. Their architecture consisted of three modified inception segments followed by convolutional and dense layers. They tested their approach on multiple datasets (CEDAR, BHSig260, UTSig) containing signatures in various scripts, achieving accuracy ranging from 80.44\% to 100\%, see Table \ref{tab1}. 

\subsection{Strengths and Weaknesses of Existing Methods}

Although existing methods demonstrate the potential of neural networks for signature verification, they mainly focus on scanned signatures rather than finger-writing on touchscreens. In addition, most approaches use complex neural architectures that may be computationally expensive for mobile devices. 
The question posed by Yang Li et al. \cite{do_we_really_need_deep_CNN_for_plant_diseases_identification} - "Do we really need a deep CNN?" - is particularly relevant in the context of mobile authentication, where computational efficiency is crucial.

\section{Methodology}\label{sec:methodology}

\subsection{Data Set}\label{AA}
Data were collected from 20 participants who each produced 2,000 finger-drawn digits (200 drawings per digit 0-9) using the index finger of their writing hand on their personal devices. The devices varied in manufacturer and operating system (Android and iOS). A control group of four participants additionally produced data on the same device to check for device bias, and one participant also wrote digits using their thumb to test the models' ability to recognize writings from different fingers of the same person. A custom-developed smartphone application was used for data collection. The application randomly selected digits for participants to draw to avoid developing drawing routines. Images were captured at a resolution of 256×256 pixels. To avoid bias from device heterogeneity, data were initially recorded in raw format as point coordinates in JSON format.

A backend system with an online interface (REST API) was developed to receive, check for inconsistencies, clean, and store the final data. The JSON data was converted to PNG image format using the OpenCV library by drawing lines between the point coordinates until a complete image was constructed. Line width could be adjusted to highlight writing characteristics, with widths of 2, 4, and 6 pixels tested. For neural network processing, images were reduced from three color channels to one, as only pixel positions were relevant. Various image sizes (32×32, 64×64, 128×128, 256×256) were tested experimentally for the Hyperparameter search. Pixel values were normalized and rounded, with filled pixels having a value of 1 and empty pixels a value of 0.

The dataset generated and analyzed during this study is available at \url{https://github.com/maanAlbalkhi/finger_trace_biometric_auth_via_nn}.

\subsection{Experimental Design}

Our experimental framework employs a personalized authentication approach, training individual models for each of the 20 participants. For binary classification architectures (shallow CNN and Mohapatra), each participant's data serves as the authorized class, while the remaining 19 participants constitute the unauthorized class. To ensure robust evaluation, unauthorized participants are distributed across training (11 participants), validation (4 participants), and test (4 participants) sets, ensuring that unknown attacker data appear only during testing. Class balance is maintained by subsampling unauthorized participant data to match the volume of authorized data. 

For both autoencoder architectures, models are trained exclusively on authorized participant data, with reconstruction thresholds set at two standard deviations above the mean reconstruction error calculated from the authorized participant's validation data. In both experimental paradigms, each authorized participant's data is partitioned using a 60:20:20 ratio for training, validation, and testing respectively.

\subsection{Training and Hyperparameter Optimization}

The hyperparameter optimization process addressed three distinct categorical subsets of parameters that collectively define the data preprocessing, model configuration and training procedure. Data-specific parameters encompassed image size and line width, while model-specific parameters included architectural elements such as the number of layers and filters, stride, padding, filter size, and activation functions. Training-specific parameters comprised optimization algorithms, learning rates, and batch size configurations. This categorization resulted in a vast hyperparameter space that required systematic exploration to identify optimal configurations.

Our optimization strategy employed a staged approach using the validation set, beginning with established best-practice values as initial parameter estimates. We then systematically explored each categorical subset sequentially, identifying the optimal parameter configuration for each subset and using these optimized values as fixed parameters when optimizing subsequent subsets. Through this iterative process across all hyperparameter categories, we determined the optimal image preprocessing and training-specific hyperparameters, ultimately yielding the model architectures described in the following subsection.

\begin{figure}[htbp]
\centerline{\includegraphics[width=\columnwidth]{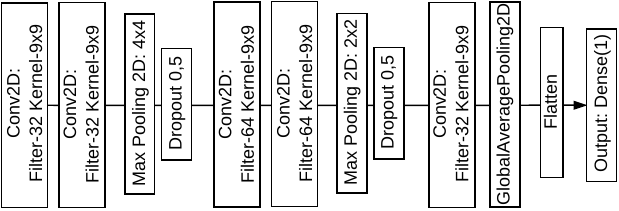}}
\caption{The architecture of the shallow CNN.}
\label{fig:arc_sh_cnn}
\end{figure}

\begin{figure}[htbp]
\centerline{\includegraphics[width=\columnwidth]{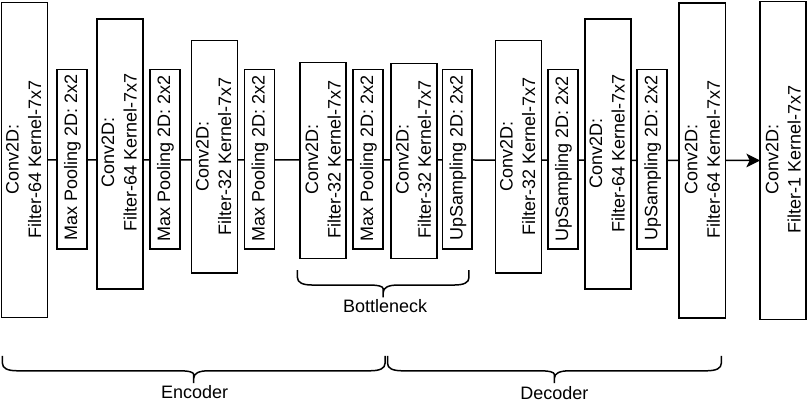}}
\caption{The architecture of the Convolutional Autoencoder.}
\label{fig:arc_cnn_ae}
\end{figure}

\begin{figure}[htbp]
\centerline{\includegraphics[width=\columnwidth]{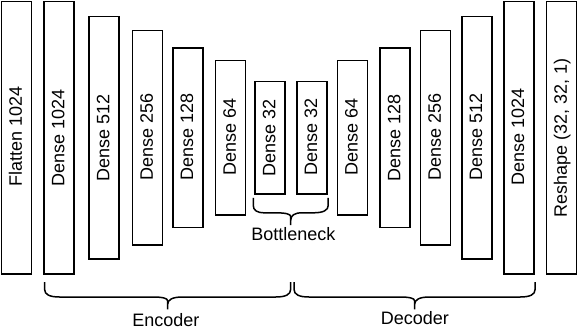}}
\caption{The architecture of the Fully Connected Autoencoder.}
\label{fig:arc_fc_ae}
\end{figure}

For binary classification architectures, datasets were balanced with a 50/50 distribution between authorized and unauthorized classes. Training incorporated early stopping mechanisms to prevent overfitting, with training halted after five consecutive epochs without improvement in validation loss as suggested in \cite{early_stopping_but_when}. For autoencoders, authentication thresholds were calculated from the mean and standard deviation of reconstruction loss values on the authorized participant's validation data, as described in the experimental design.

\subsection{Neural Network Architectures}
Four distinct neural network architectures were implemented and evaluated for finger-drawn symbol authentication, encompassing both discriminative and generative approaches.

\begin{itemize}
    \item \textbf{Mohapatra's Modified Inception-V1 Architecture \cite{8985925}:} This complex CNN architecture, originally designed for signature verification, was adapted for finger writing authentication. The network consists of three inception segments followed by convolutional and dense layers, providing multi-scale feature extraction capabilities suitable for biometric pattern recognition.
    \item \textbf{Shallow CNN Architecture:}  A proposed lightweight CNN designed to reduce computational requirements while maintaining authentication performance. Following hyperparameter optimization, the final architecture comprises two convolutional layers followed by a MaxPooling layer (4×4), two additional convolutional layers followed by a MaxPooling layer (2×2), another convolutional layer, a GlobalAveragePooling layer, and a final dense layer for binary classification. See Figure \ref{fig:arc_sh_cnn} for an illustration of this architecture.
    \item \textbf{Convolutional Autoencoder:} This architecture employs anomaly detection principles for authentication, treating unauthorized access attempts as anomalous patterns. The encoder consists of four convolutional layers with filter configurations of 64-64-32-32, while the decoder mirrors this structure with 32-32-64-64 filters, reconstructing the input image for threshold-based authentication decisions. Compare Figure \ref{fig:arc_cnn_ae} for an illustration of this architecture.
    \item \textbf{Fully Connected Autoencoder:} This dense layer-based architecture features six layers in both encoding and decoding phases. Input images are flattened before processing and reshaped to their original dimensions at output. The bottleneck layer compresses the input representation to a 32-dimensional vector, enabling efficient anomaly detection through reconstruction error analysis. As can be seen in Figure \ref{fig:arc_fc_ae}.
\end{itemize}

\subsection{Evaluation Methods}\label{SCM}
Model performance was evaluated using several metrics that Elisa Bertino et al. \cite{computer_security_ESORICS_2021} identified at the Computer Security Conference (ESORICS 2021) as important metrics for evaluating authentication systems:
\begin{itemize}
    \item \textbf{False Acceptance Rate (FAR):} The probability that the system incorrectly allows access to an unauthorized person. This measures security - a high FAR indicates the system is too lenient and compromises security. Which is equivalent to the False Negative Rate (FNR) and hence 1-True Positive Rate (TPR).
    \begin{equation} FAR = \frac{FP}{FP + TN}\end{equation}
    \item \textbf{False Rejection Rate (FRR):} The probability that the system incorrectly rejects a legitimate user. This affects user experience - a high FRR means genuine users are frequently denied access. Also known as False Positive Rate (FPR).
    \begin{equation} FRR = \frac{FN}{FN + TP}\end{equation}
    \item \textbf{Equal Error Rate (EER):} The point where FAR and FRR are equal, providing a single value that balances security and usability concerns. This is useful for comparing different authentication systems.
    \begin{equation} EER = \frac{FAR + FRR}{2}\end{equation}
    \item \textbf{Accuracy (ACC):} The ratio of correct predictions to total predictions, providing an overall measure of system performance.
    \begin{equation} ACC = \frac{TP + TN}{TP + FP + TN +FN}\end{equation}
\end{itemize}
Where TP = True Positives, FP = False Positives, TN = True Negatives and FN = False Negatives.

Receiver Operating Characteristic (ROC) curves were also constructed to determine appropriate threshold values for each model and participant. Final results represent performance metrics averaged across all 20 individual participant models.

\section{Results}\label{sec:results}

This section presents the authentication performance results for four neural network architectures evaluated on finger-drawn digit authentication. Individual models were trained for each of the 20 participants, with results averaged across all participant models. The evaluation encompasses both discriminative approaches (Shallow CNN and Mohapatra's Modified Inception-V1) and generative approaches (Convolutional and Fully Connected Autoencoders).

\subsection{Overall Performance Comparison}

Table \ref{tab2} shows that CNN architectures significantly outperformed autoencoder approaches. Both CNN methods achieved $\sim$89\% accuracy with AUC values above 0.95, while autoencoders reached only 77\% and 72.2\% accuracy. CNN architectures demonstrated balanced error rates with FAR in the range 12\%-16\% and FRR below 9\%, whereas autoencoders showed higher and more variable error rates, particularly the FC-Autoencoder with 30.5\% FRR.

\subsection{CNN Performance Analysis}
Figures \ref{fig:sh_cnn_roc} and \ref{fig:inc_cnn_roc} illustrate the ROC curves and discrimination threshold plots for the Shallow CNN and Mohapatra's Modified Inception-V1 architectures respectively. Both architectures demonstrate excellent discriminative capability with nearly identical AUC values of 0.9515 and 0.9562. The ROC curves show steep initial rises and minimal deviation from the ideal top-left corner, indicating strong separation between authorized and unauthorized samples. The discrimination threshold plots reveal well-separated class distributions, with unauthorized samples (Class 0) concentrated at lower loss values and authorized samples (Class 1) at higher loss values, demonstrating clear decision boundaries. Figure \ref{fig:frr_far} demonstrates the trade-off between FAR and FRR across the full range of decision thresholds for the Shallow CNN.

Performance comparison between the two CNN architectures reveals minimal differences in authentication accuracy, with the Shallow CNN achieving 89\% accuracy versus 88.6\% for the Mohapatra architecture. However, the Shallow CNN offers significant computational advantages, requiring less memory and inference time compared to the complex Inception-based model. This makes the Shallow CNN particularly suitable for resource-constrained mobile deployment scenarios while maintaining comparable authentication performance.

\subsection{Autoencoder Performance Analysis}
\begin{figure}[htbp]
\centerline{\includegraphics[width=\columnwidth]{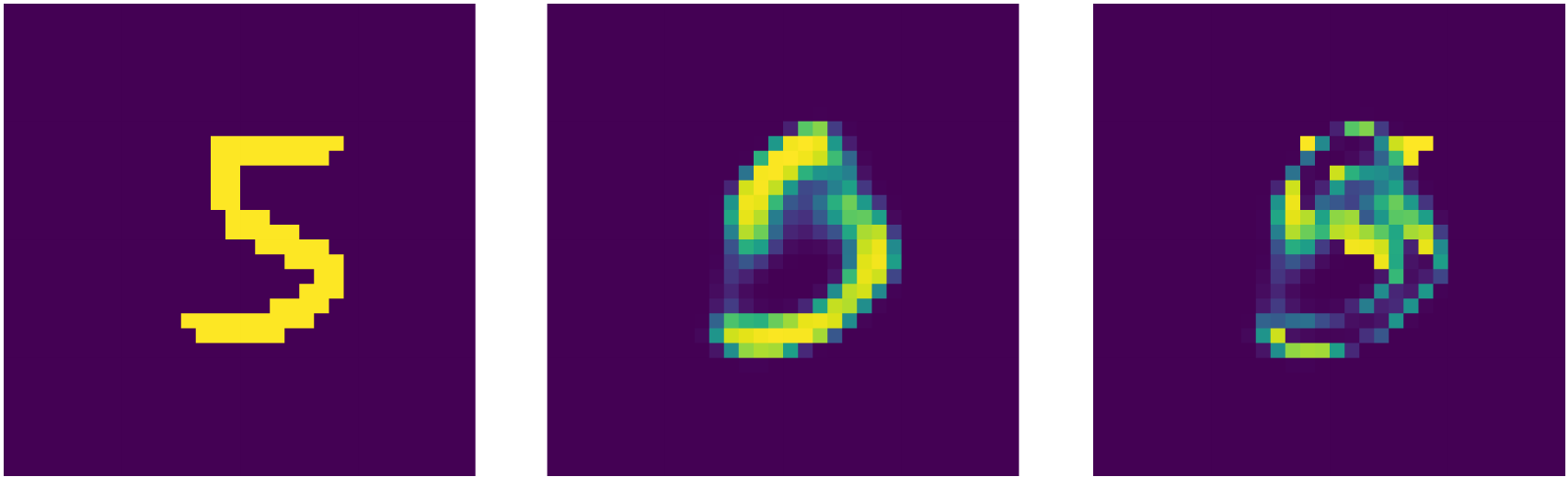}}
\caption{Anomaly detection using the CNN Autoencoder for an unauthorized sample. The images illustrate the input, reconstruction, and reconstruction error from left to right.}
\label{fig:ae_example}
\end{figure}

The CNN-Autoencoder outperformed the FC-Autoencoder (77\% vs 72.2\% accuracy), suggesting convolutional features better capture spatial patterns in finger-drawn digits, as shown in Table \ref{tab2}. Figure \ref{fig:ae_example} illustrates the anomaly detection mechanism: the unauthorized '5' sample reconstructs as a '6' or '0', indicating it shares more similarity with genuine '6'/'0' patterns than genuine '5' patterns. This reconstruction error enables authentication through anomaly detection.

\subsection{Cross-Architecture Insights}
The performance gap between CNN and autoencoder approaches highlights the effectiveness of discriminative versus generative methods for this biometric task. CNNs benefit from direct class separation optimization, while autoencoders rely on indirect authentication through reconstruction quality. The consistent CNN performance suggests lightweight architectures are sufficient for finger-drawn authentication, with the Shallow CNN providing optimal balance between accuracy and computational efficiency.

\begin{figure}[htbp]
\centerline{\includegraphics[width=\columnwidth]{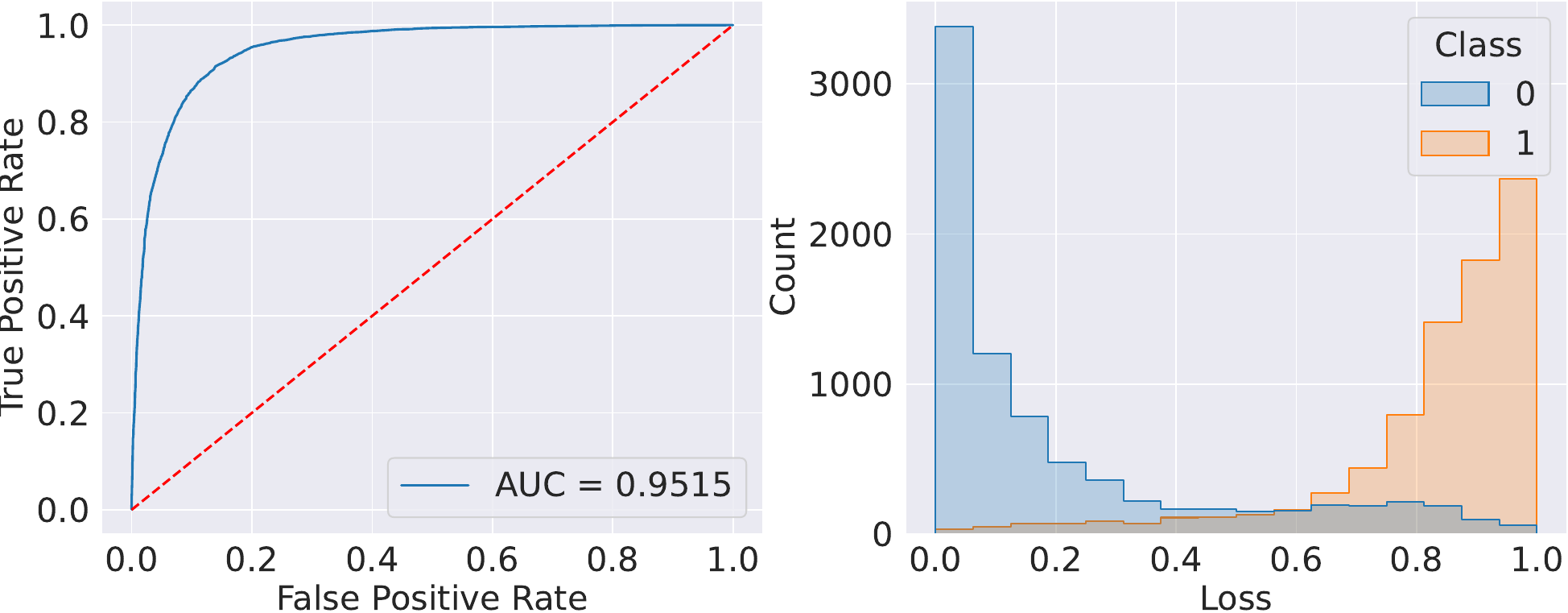}}
\caption{Averaged ROC (left) and Discrimination Threshold Plot (right) over all participants for the shallow CNN.}
\label{fig:sh_cnn_roc}
\end{figure}

\begin{figure}[htbp]
\centerline{\includegraphics[width=\columnwidth]{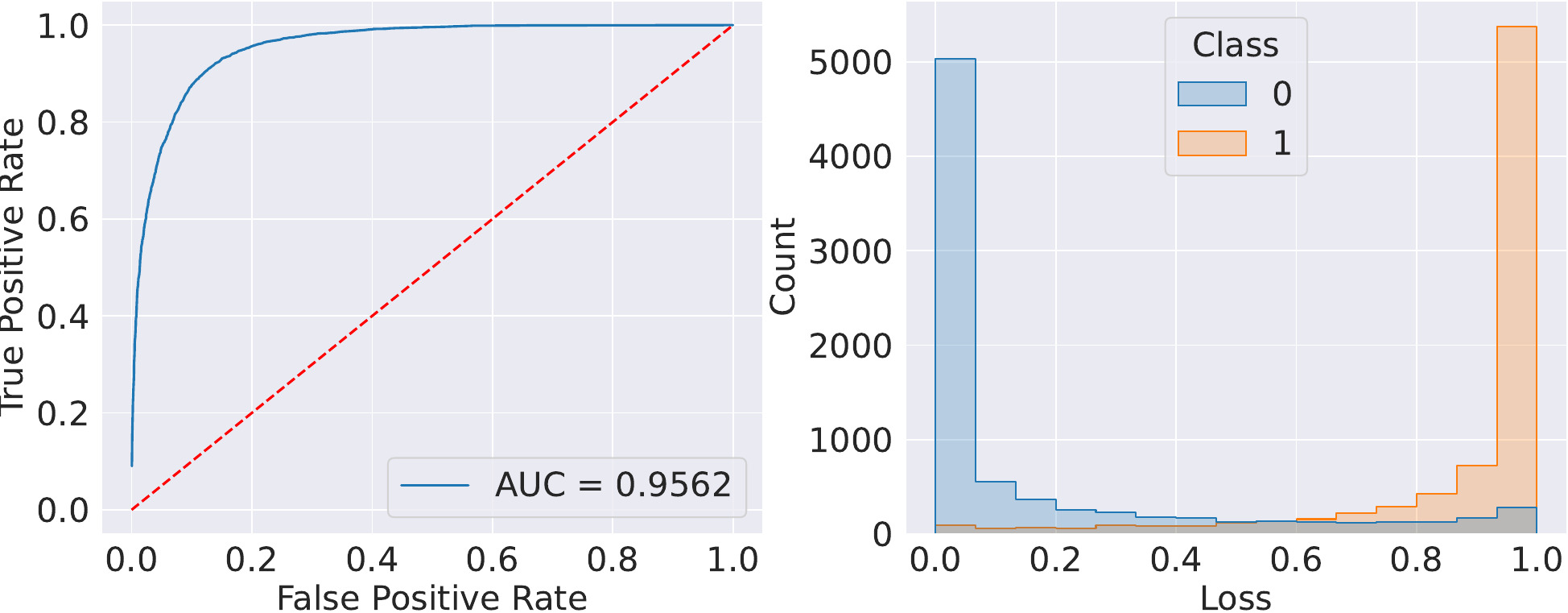}}
\caption{Averaged ROC (left) and Discrimination Threshold Plot (right) over all participants for Mohapatra's CNN.}
\label{fig:inc_cnn_roc}
\end{figure}

\begin{figure}[htbp]
\centerline{\includegraphics[width=\columnwidth]{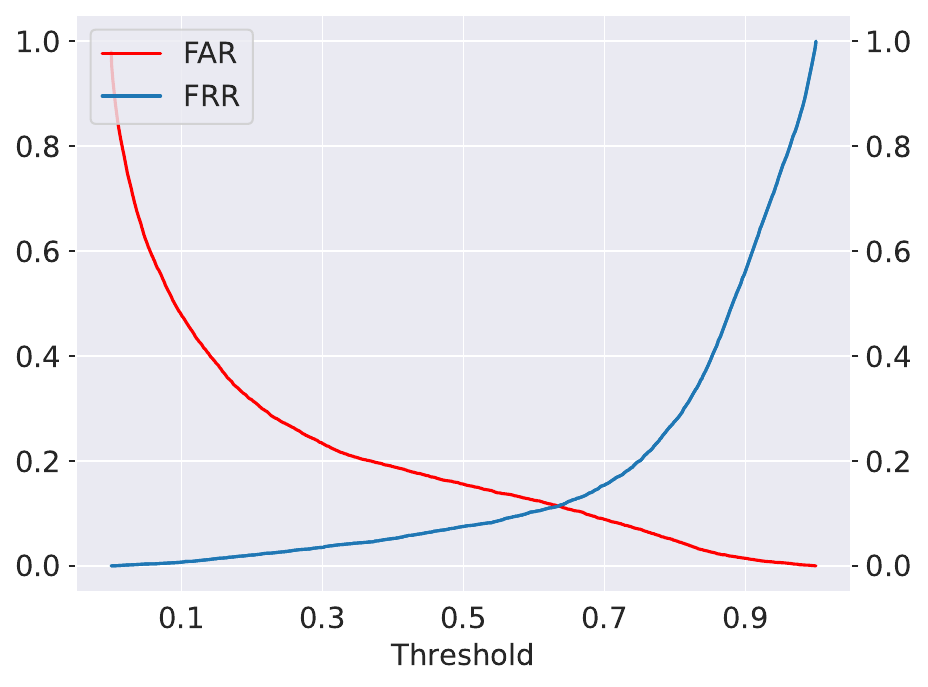}}
\caption{FAR and FRR trade-off curves for the Shallow CNN across different decision thresholds. The intersection point indicates the Equal Error Rate (EER) at 11.5\%.}
\label{fig:frr_far}
\end{figure}

\begin{figure}[htbp]
\centerline{\includegraphics[width=\columnwidth]{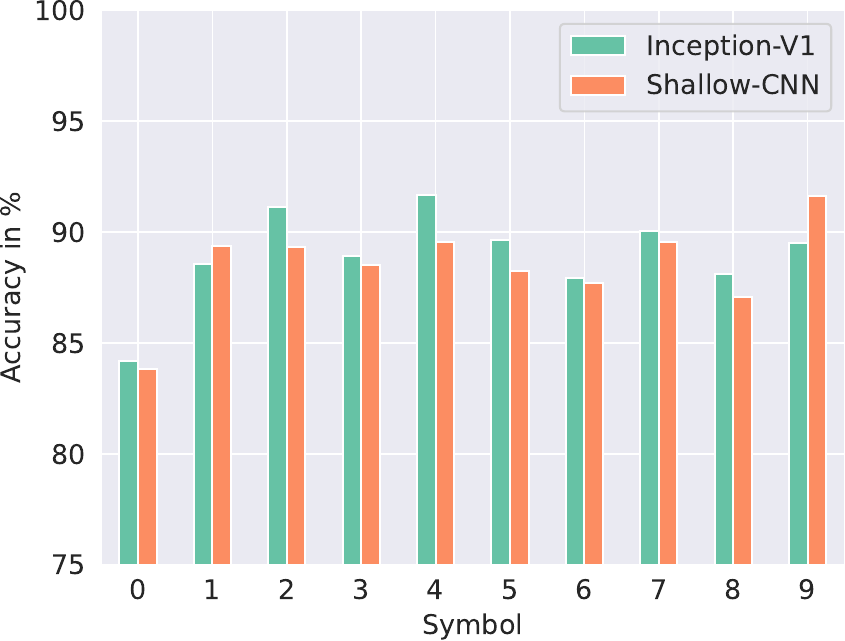}}
\caption{Shallow CNN and Mohapatra's modified inception-v1 accuracy when tested on each digit class individually.}
\label{fig:digits}
\end{figure}

\begin{figure}[htbp]
\centerline{\includegraphics[width=\columnwidth]{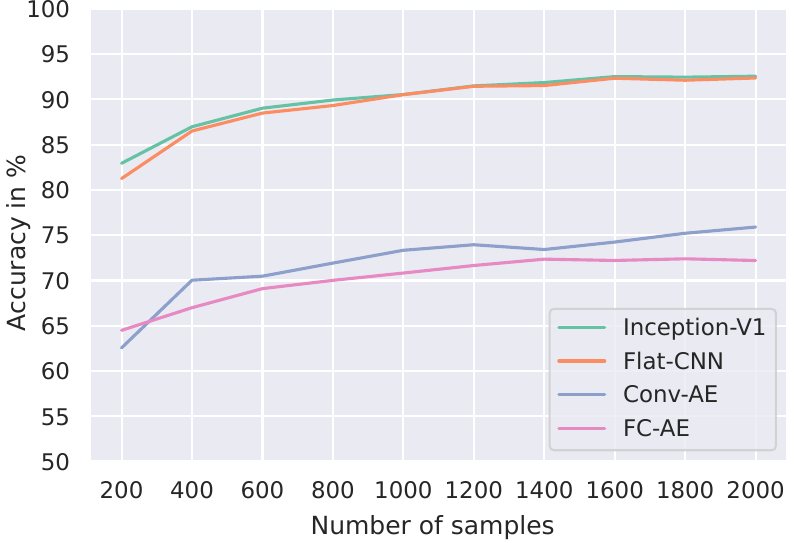}}
\caption{Influence of the number of training samples on the resulting accuracy for each architecture.}
\label{fig:influence_data_amount}
\end{figure}

\begin{table}[htbp]
\caption{Achieved results for each architecture, averaged over all participants}
\begin{center}
\begin{tabular}{|c|c|c|c|c|c|}
\hline
\textbf{Architecture} & \textbf{FAR} & \textbf{FRR} & \textbf{EER} & \textbf{ACC} & \textbf{AUC}\\
\hline
 Shallow CNN & 15.6\% & 7.5\% & 11.5\% & 89.1\% & 0.95\\
 \hline
 Mohapatra et al. & 12.8\%  & 8.9\% & 10.9\% & 88.5\% & 0.96\\ 
 \hline
 CNN-Autoencoder & 27.9\% & 20.3\% & 24.1\% & 77\% & 0.83 \\ 
 \hline
 FC-Autoencoder & 25.1\% & 30.5\% & 27.8\% & 72.2\% & 0.8 \\
\hline
\end{tabular}
\label{tab2}
\end{center}
\end{table}

\subsection{Impact of Symbol Complexity and Data Quantity}\label{sub:impact_of_symbol}
Figure \ref{fig:digits} illustrates the performance of the trained Shallow CNN and Mohapatra architecture when tested separately on each digit class. It can be seen that more complex symbols like '4' and '9' achieved higher accuracy across all models, while simpler shapes like '0' and '8' yielded lower accuracy. This suggests that complex structures with multiple strokes or direction changes provide better recognition rates.
Figure \ref{fig:influence_data_amount} shows the influence the quantity of training data has on the overall model accuracy. The CNN models showed no improvement in accuracy beyond 1,600 training samples, while the Fully Connected Autoencoder plateaued at 1,400 training samples. The Convolutional Autoencoder continued to improve with increasing data volume.

\section{Discussion}\label{sec:discussion}

\subsection{Interpretation of Results}

Our results demonstrate the feasibility of person authentication using finger-drawn simple symbols on touchscreens, with convolutional neural networks (CNNs) achieving up to 89\% accuracy. Shallow CNNs performed comparably to the deeper Mohapatra architecture despite having fewer parameters (749,345 vs. 1,933,729), supporting previous findings by Yang Li et al. that deeper architectures may be unnecessary for constrained classification tasks \cite{do_we_really_need_deep_CNN_for_plant_diseases_identification}.

The hyperparameter optimization step showed that input resolution and line width played a key role: smaller images (32×32 or 64×64) with thicker lines (6 pixels) outperformed larger, sparser inputs. This suggests that concentrating relevant stroke information in a smaller area helps the model focus on signal over noise, improving feature extraction and generalization. Larger convolutional kernels also proved beneficial, likely due to their ability to capture global stroke patterns.

Autoencoder-based models, while less accurate ($\sim$75\%), showed potential for anomaly detection–based authentication. However, their weaker performance relative to CNNs (a $\sim$14\% gap) indicates that further refinement is needed to make them viable for standalone use.

The authentication system allows for tuning the trade-off between security and usability by adjusting the decision threshold. In our binary classification approach, adjusting the threshold allows system designers to tune the operating point along the FAR-FRR curve to meet specific application requirements \cite{jain2004introduction}. Figure \ref{fig:frr_far} illustrates this trade-off for the Shallow CNN, showing how FAR decreases and FRR increases as the threshold becomes more restrictive.

For security-critical applications like banking or healthcare, operating at a lower FAR is essential even if legitimate users must occasionally retry authentication. As shown in Figure \ref{fig:frr_far}, the threshold can be adjusted to achieve FAR below 5\% at the cost of FRR increasing to approximately 25\%. While this represents a usability trade-off—requiring legitimate users to attempt authentication multiple times, the enhanced security may be necessary for high-risk scenarios. For convenience-focused applications like content unlocking or personalization, a lower FRR may be prioritized. The ROC curves in Figures \ref{fig:sh_cnn_roc} and \ref{fig:inc_cnn_roc} demonstrate that our models provide flexibility for threshold adjustment across this spectrum.

This usability drawback can be partially mitigated through two complementary strategies. First, encouraging the use of more complex symbols: according to our analysis in Section \ref{sub:impact_of_symbol}, complex symbols are inherently more distinguishable and lead to better authentication performance. Second, authentication strength can be significantly enhanced through sequential symbol inputs. Assuming independent authentication attempts, requiring multiple correctly authenticated symbols exponentially reduces the effective FAR. For instance, two-symbol authentication reduces FAR from 15.6\% to approximately 2.4\% (0.156²), while three-symbol sequences achieve FAR of approximately 0.38\% (0.156³). This multi-symbol approach, like PIN length in traditional authentication, provides a tunable parameter to raise the system's security to levels suitable for high-stakes applications, while maintaining architectural simplicity.

\subsection{Security Considerations and Impersonation Resistance}

The current implementation demonstrates authentication based on spatial characteristics of finger-drawn symbols. While the 89\% accuracy and 15.6\% FAR represent competitive performance for single-factor authentication, several strategies can enhance security against sophisticated impersonation attacks:

\begin{itemize}
    \item \textbf{Secret symbol patterns:} Unlike the standardized digits (0-9) used in this study for experimental consistency, real-world deployments should employ user-defined secret patterns. Similar to Android's pattern lock, users could draw arbitrary shapes or symbols known only to them, eliminating the risk of attackers studying standardized digit-drawing styles.
    \item \textbf{Multi-symbol authentication:} Requiring authentication across multiple sequential symbols (e.g., a 3-digit sequence) exponentially reduces FAR. With single-symbol FAR of 15.6\%, a three-symbol sequence achieves effective FAR of $\sim$0.38\%.
    \item \textbf{Integration with dynamic features:} Our data collection included temporal information (drawing speed, stroke order, pressure) that were not utilized in the current analysis. Future work incorporating these behavioral dynamics—such as drawing velocity profiles, pressure patterns, and device grip characteristics could significantly enhance discriminative power and impersonation resistance. These features are particularly valuable as they capture unconscious behavioral patterns difficult for attackers to replicate even with visual access to the drawn symbol.
    \item \textbf{Hybrid authentication schemes:} Finger-drawn symbols are most effective as one factor in multi-factor authentication (MFA) systems. When combined with device possession (SIM card, trusted device), knowledge factors (PIN), or traditional biometrics (fingerprint, facial scan), the compound security significantly exceeds individual factor limitations.
\end{itemize}

\subsection{Comparison to Related Work}

Compared to existing literature, our method achieves competitive results using a simple and efficient architecture. Volaka et al. reported 88\% accuracy on touchscreen interaction patterns, while our CNN models reached 89\% using finger-drawn digits\cite{volaka2019towards}. Takahashi et al. achieved a 10.6\% Equal Error Rate (EER) with handcrafted features, whereas our CNNs reduced the EER to as low as 4.92\% \cite{takahashi2021smartphone}.

Mohapatra et al. reported near-perfect performance in signature verification, but their system relied on scanned, high-resolution handwritten signatures, a modality less suitable for real-time mobile interaction \cite{8985925}. In contrast, our method is designed for direct finger input and optimized for mobile deployment, offering both high accuracy and practical usability.

\subsection{Applications and Deployment Potential}

Finger-drawn symbol authentication offers promising applications in various security contexts:

\begin{itemize}
    \item \textbf{Primary authentication}: It can replace PINs or passwords on mobile devices with a more intuitive input modality.
    \item \textbf{Multi-factor authentication (MFA)}: It complements existing methods by combining biometric (stroke dynamics) and knowledge-based (symbol shape) factors.
    \item \textbf{Continuous authentication}: The non-intrusive nature of finger-drawn inputs makes the method suitable for periodic re-authentication during usage.
    \item \textbf{Granular access control}: It allows selective protection of sensitive applications (e.g., banking, secure messaging), while maintaining default access for less critical features.
\end{itemize}

Furthermore, similar to findings by Shixuan Wang et al. on motion-sensor-based authentication, future systems could enhance robustness by adapting to varying device orientations or usage contexts \cite{motion_sensor_based_continuous_authentication}. Our method could similarly benefit from incorporating contextual factors such as device orientation.

\subsection{Limitations}

Despite strong performance, several limitations warrant consideration:

\begin{itemize}
    \item \textbf{Symbol simplicity}: The system showed reduced accuracy for visually simple digits (e.g., “1”), likely due to lower intra-user variation and fewer distinguishing features.
    \item \textbf{Device heterogeneity}: Differences in screen size, resolution, and touch sensitivity across devices may influence writing behavior and affect generalization.
    \item \textbf{Dataset scope}: The study focused exclusively on digits (0–9), limiting generalizability to arbitrary or user-defined symbols.
    \item \textbf{Spatial-only feature extraction}: While our data collection captured rich behavioral dynamics including drawing velocity, the current analysis utilized only spatial characteristics. This represents a significant opportunity for future work, as behavioral biometrics research demonstrates that dynamic features (stroke velocity, pressure patterns, drawing order) often provide stronger discriminative power than spatial features alone.
    \item \textbf{Autoencoder underperformance}: The large performance gap between CNNs and autoencoders suggests that current anomaly detection methods are not yet competitive, though the approach remains promising.
\end{itemize}

\subsection{Conclusion and Outlook}

This study validates finger-drawn symbol authentication as a viable and accurate biometric method for touchscreen devices. A shallow CNN model achieved high accuracy with a small parameter count, making it ideal for mobile deployment where computational efficiency is critical. The model was robust to device variability and capable of distinguishing not just between users, but also between different fingers of the same user, indicating it captured fine-grained, individual writing characteristics.

Autoencoder models, while currently less accurate, offer a foundation for anomaly-based authentication and may become more viable with architectural or training improvements.

Several promising directions emerge for future research:

\begin{itemize}
    \item Incorporating dynamic behavioral features, i.e. temporal data (drawing velocity, acceleration profiles, stroke order) and device motion signals (accelerometer, gyroscope) to enhance authentication accuracy and impersonation resistance. These unconscious behavioral patterns are particularly valuable for continuous authentication and difficult for attackers to replicate.
    \item Conducting user studies to empirically determine acceptable FAR/FRR trade-offs across different application contexts (e.g., banking vs. general access), and implementing adaptive threshold mechanisms that adjust based on risk context.
    \item Expanding symbol sets beyond digits to include letters, shapes, or arbitrary user-defined patterns
    \item Testing the shallow CNN on physical mobile devices to assess real-world latency and energy efficiency
    \item Exploring hybrid models combining CNN classification and anomaly detection
    \item Developing adaptive methods that accommodate user drift or changing input behavior over time
    \item Improving autoencoder architectures to close the performance gap with CNNs
    \item Enabling authentication via arbitrary drawn patterns for enhanced security
    \item Investigating adaptive threshold tuning strategies to balance security and usability dynamically, e.g.  lowering the False Acceptance Rate (FAR) for sensitive scenarios at the cost of a higher False Rejection Rate (FRR), which could be partially offset by encouraging more complex symbol use
\end{itemize}

Overall, finger-drawn symbols provide a secure, user-friendly alternative or complement to traditional authentication methods. With further research, such systems could offer fast, low-friction, and personalized security mechanisms for modern touchscreen devices.

\section*{Acknowledgment}

The project on which this report is based was funded by the Federal Ministry of Research, Technology and Space under the funding code “KI-Servicezentrum Berlin-Brandenburg” 16IS22092. Responsibility for the content of this publication remains with the author. We thank Lutz Reiter for his expertise and assistance in developing the application used for data collection in our experiments.

\printbibliography
\end{document}